\documentclass[11pt,a4paper]{article}
\usepackage[hyperref]{emnlp2020}
\usepackage{times}
\usepackage{latexsym}

\usepackage{microtype}

\aclfinalcopy % Uncomment this line for the final submission
\title{Machines Getting with the Program:\\ Understanding Intent Arguments of Non-Canonical Directives}

\author{Won Ik Cho\textsuperscript{1}, Young Ki Moon\textsuperscript{2,3}*, Sangwhan Moon\textsuperscript{4,5}*, Seok Min Kim\textsuperscript{1}, Nam Soo Kim\textsuperscript{1}\\
  Department of Electrical and Computer Engineering and INMC, Seoul National University\textsuperscript{1},\\
	Department of Computer Engineering, Inha University\textsuperscript{2}, Voithru Co., Ltd.\textsuperscript{3},\\
	 Department of Computer Science, Tokyo Institute of Technology\textsuperscript{4}, Odd Concepts Inc.\textsuperscript{5}\\
  \texttt{wicho@hi.snu.ac.kr, ykmoon0814@gmail.com,  sangwhan@iki.fi}\\ \texttt{smkim@hi.snu.ac.kr, nkim@snu.ac.kr} \\}

\date{}

\usepackage{graphicx}
\usepackage{tabularx}
\usepackage{soul}
\usepackage{kotex}
\usepackage{multirow}
\usepackage[normalem]{ulem}
\usepackage{enumitem}
\usepackage{marvosym}

\usepackage{epstopdf}

\usepackage{hyperref}
\usepackage{xstring}

\usepackage{lipsum}

\makeatletter
\def\blfootnotedollar{\gdef\@thefnmark{**}\@footnotetext}
\makeatother

\makeatletter
\def\blfootnotestar{\gdef\@thefnmark{*}\@footnotetext}
\makeatother

\begin{document}
\maketitle
\begin{abstract}
Modern dialog managers face the challenge of having to fulfill human-level conversational skills as part of common user expectations, including but not limited to discourse with no clear objective. Along with these requirements, agents are expected to extrapolate intent from the user’s dialogue even when subjected to non-canonical forms of speech. This depends on the agent’s comprehension of paraphrased forms of such utterances. Especially in low-resource languages, the lack of data is a bottleneck that prevents advancements of the comprehension performance for these types of agents. In this regard, here we demonstrate the necessity of extracting the intent argument of non-canonical directives in a natural language format, which may yield more accurate parsing, and suggest guidelines for building a parallel corpus for this purpose. Following the guidelines, we construct a Korean corpus of 50K instances of question/command-intent pairs, including the labels for classification of the utterance type. We also propose a method for mitigating class imbalance, demonstrating the potential applications of the corpus generation method and its multilingual extensibility.
\end{abstract}

%\begin{document}

%\maketitleabstract

\blfootnotestar{Both authors contributed equally to this manuscript.}

\section{Introduction}  

The advent of smart agents such as Amazon Echo and Google Home has shown relatively wide market adoption. Users have been familiarized with formulating questions and orders in a way that these agents can easily comprehend and take actions. Given this trend, particularly for cases where questions have various forms such as \textit{yes/no}, \textit{alternative}, \textit{wh-}, \textit{echo} and \textit{embedded} \cite{huddleston1994contrast}, a number of analysis techniques have been studied in the domain of semantic role labeling \cite{shen2007document} and entity recognition \cite{molla2006named}. Nowadays, various question answering tasks have been proposed \cite{yang2015wikiqa,rajpurkar2016squad} and have yielded systems that demonstrate significant advances in performance. Studies on the parsing of canonical imperatives \cite{matuszek2013learning} have also been done for many 
intelligent agents.

\begin{figure}
	\centering
	\includegraphics[width=0.5\textwidth]{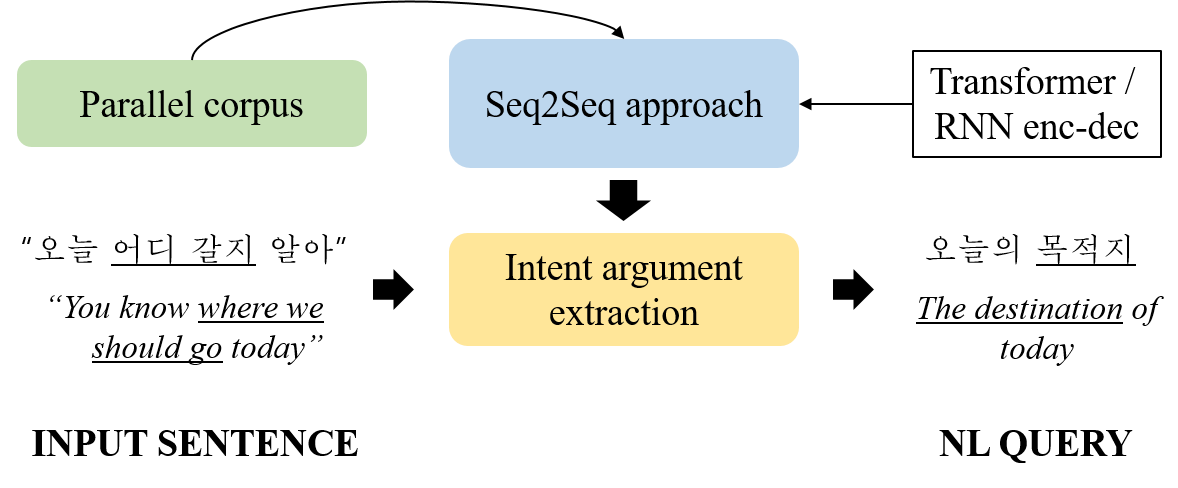}
	\caption{A diagram of the proposed extraction scheme. Unlike in the Korean example that is to be investigated, in English translation, the \textit{wh-}related noun (here, \textit{destination}) is placed at the head part of the output.} \label{fig:fig2}
\end{figure} 

However, discerning the intention from a conversational and non-canonical directive (question or command) and correctly extracting its intent argument is still a challenge. It usually matters when the user is not familiar with the \textit{canonical} commands, namely where the direct speech act meets the genuine intention. %; e.g., \textit{wh-}first questions with pure \textit{wh-}questions. 
That is, sometimes, the speech act can be hard to guess merely from the sentence form, as in inferring (1),\smallskip\\
(1) \textit{why don’t you just call the police}\smallskip\\
as a representation of the to-do list ‘\textit{to call the police}’. Although advanced dialog managing systems may generate a plausible reaction regarding the input utterance, it is different from extracting the exact intent argument (a \textit{question set} or a \textit{to-do-list}) that should be investigated for actual operation.

Additional complexity is introduced when the target text is in a speech recognition context, as the input text may not contain punctuation. For example, given an unclear declarative question \cite{gunlogson2002declarative} such as (2),\smallskip\\
(2) \textit{you know where we should go today}\smallskip\\
a human listener can interpret the subject of inquiry as ‘\textit{the destination of today}’, while this can be challenging for a machine. The basis of our work is that if a system is trained to extract a structured natural language (NL) query from directive sentences, it may help the language understanding systems be more robust at understanding non-canonical expressions in executing the command.

%\blfootnotedollar{The authors appreciate Siyeon Natalie Park for having discussion and suggesting the great idea for the title.}

Some may argue that the structured information retrieval we aim to support may benefit from the data augmentation technologies that are concurrent with the studies on paraphrasing \cite{xie2019unsupervised,kumar2020data}. However, complexities as in the examples above have not seen much exploration outside of English, especially in the context of languages with a distinguished syntax or cases which do not use Latin-like alphabets. Also, it is not guaranteed that such technologies fit with less explored languages, where sufficient pre-training resources may not be 
readily available.

As a more concrete example, in the Korean language, the morphology is agglutinative, the syntax is head-final, and scrambling (non-deterministic permutations of word/phrase ordering) is a common practice between native speakers. Primarily, the agglutinative property of Korean requires additional morphological analysis, which makes it challenging to identify the component of the sentence that has the most substantial connection to core intent. Moreover, the head-finality characteristic introduces an additional layer of complexity, where an under-specified sentence ender incorporates a prosodic cue which requires disambiguation to comprehend the original intent \cite{yun2019meaning, cho2019disambiguating}. Finally, considering the scrambling aspect, which frequently happens in spoken utterances, further analysis is required on top of recognizing the entities and extracting the relevant phrases. These make it difficult for dialog managers to directly apply conventional analysis methods that have been used in Germanic or other Indo-European languages.

In this paper, based on such aspects of the conversation-style utterances of Korean, we propose a structured NL query\footnote{Hereafter, we interchangeably use \textit{NL query} and \textit{(intent) argument} to indicate the structured core content, depending on the context.} extraction scheme, which can help enrich the human-like conversation with artificial intelligence (AI). For automation, we construct a corpus of sentence-phrase pairs via annotation and then augment the dataset to mitigate class imbalance, demonstrating the flexibility, practicality, and extensibility of the proposed methods. To further prove that the scheme is not limited to a specific language, we demonstrate the methodology using English examples and supplement specific cases with Korean. We describe the followings as our contribution to the field:
\begin{itemize}
    \item We propose the scheme for building the parallel corpora of non-canonical Korean directives and their intent arguments, along with speech act type labeled, and release it publicly.
    \item We suggest a visible result on the content extraction scheme with conventional Seq2Seq systems, probing the application potential.
\end{itemize}

\section{Concept and Related Work}

The theoretical background of this proposal builds on literature from speech act \cite{searle1976classification} and formal semantics \cite{portner2004semantics}. Although many task-oriented systems identify the intents as a specific action that the agent should take \cite{liu2016attention,li2018microsoft}, to make our approach generic in the aspect of the domain and sentence structures, we hypothesized that it would be beneficial for the natural language understanding (NLU) modules first to recognize the directiveness and represent the core content in a structured format. 

% We believe that the closest problem we have to this task is formulating a question set (QS) or to-do-list (TDL) with multiple possible utterance permutations (Table 1) \cite{portner2004semantics}. While these concepts have stronger relations with the domain of syntactic properties, we extend on this to speech act level to reflect common patterns in a human dialog form.

% \begin{table}[h]
% 	%\footnotesize
% 	\centering
% 	%\resizebox{\columnwidth}
% 	\resizebox{\columnwidth}{!}{%
% 		\begin{tabular}{|c|c|c|c|}%{0.7\textwidth}
% 			\hline
% 			\textbf{Type}           & \textbf{Denotations} & \textbf{Discourse Component} & \textbf{Force} \\ \hline
% 			\textbf{Declaratives}   & proposition (p)         & Common Ground                & Assertion      \\ \hline
% 			\textbf{Interrogatives} & set of propositions (q) & Question Set                 & Asking         \\ \hline
% 			\textbf{Imperatives}    & property (P)            & To-Do List Function          & Requiring      \\ \hline
% 		\end{tabular}%
% 	}
% 	\caption{Clause types and their properties (Portner, 2004).}
% 	\label{my-label}
% \end{table}

Once an utterance is identified to be directive, conventional systems rely on slot-filling to extract the item and argument \cite{li2018microsoft,haghani2018audio}, where the number of the categories is generally restricted. Instead, for non-task-oriented dialogues, we hypothesized that the arguments should be attained in NL format rather than structured data, by, e.g., rewriting the utterances into some nominalized or simplified terms which correspond to the source text. There have been studies on paraphrasing of questions concerning the core content \cite{dong2017learning}, but little has been done on NL formalization. Also, our study targets the extraction of commands, which is equivalently essential but has not been widely explored outside of the robotics domain \cite{matuszek2010following,matuszek2013learning}.

The closest problem to this task is probably semantic parsing \cite{berant2014semantic,su2017cross} and structured query language (SQL) generation, \citet{zhong2017seq2sql} which propose Seq2Seq \cite{sutskever2014sequence}-like architectures to transform NL input into a structured format. These approaches provide the core content of the directive utterances as a sequence of queries, both utilizing it in paraphrasing \cite{berant2014semantic} or code generation \cite{zhong2017seq2sql}. However, the proposed source sentence formats are usually \textit{canonical} and mostly information-seeking, rather than being in a colloquial context.

Our motivation builds on the basis that real-world utterances as input (e.g., smart home commands from the less tech-familiar audience), in particular for Korean, can diverge from the expected input form, to the \textit{non-canonical} utterances that require actual comprehension for classifying as a question or command. Moreover, as we discuss in the latter part of our work, we intend the extracted NL queries to be reusable as building blocks for efficient paraphrasing, following the approach in \newcite{berant2014semantic}. 

Recently, in a related view, or stronger linguistic context emphasis, guidelines for identifying non-canonical natural language questions or commands have been suggested for Korean \cite{cho2018speech}. We build on top of this corpus for the initial dataset creation, and extend the dataset with additional human-generated sentences. %However, the domain of the sentences are restrained and the compatibility has not been shown for the other languages with different syntax and lexicon.

\begin{figure*}
	\centering
	\hspace*{+0.3in}
	\includegraphics[width=0.85\textwidth]{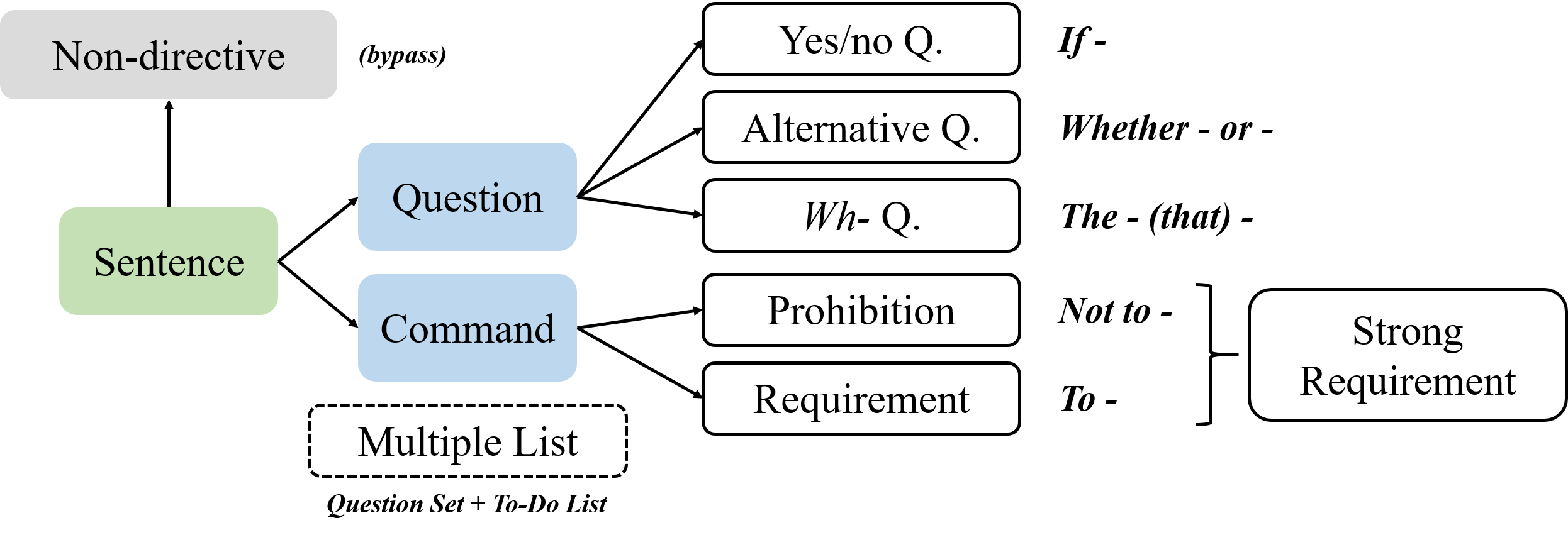}
	\caption{A simple description on the categorization and annotation. The sentence is either a given text utterance or a transcript. The lexicons on the right side denote the head of the arguments (which goes to the tail of a phrase in Korean). Multiple list denotes the rare cases where question and command co-exist, but was not detected in the construction phase. The strong requirement, which is a serial placement of PH and REQ, is to be explained afterward since it originates in an empirical study and may not be a universal phenomenon.} \label{fig:fig2}
\end{figure*} 

\section{Proposed Scheme}

In this section, we describe the corpus construction scheme along with the motivation of this work. As discussed in the first section, our goal is to propose a guideline for discerning the intent argument for conversational and non-canonical questions and commands. These appear a lot in everyday life, but unlike cases where the input is in a canonical form, algorithmically extracting the core intent is not straightforward. We suggest that a data-driven methodology should be introduced for this task, which can be done by creating a corpus annotated with the core content of the utterances. While our work in this paper is for Korean, the example sentences and the proposed structured scheme are provided in English, for demonstrative purposes.

\subsection{Identifying Directives}

% FIXME: Revisit this part later

Identifying directive utterances is a fundamental part of this work, though our main content is not just classification. Thus, at this moment, we briefly demonstrate the Korean corpus whose guideline is for distinguishing such utterances from the non-directives such as fragments and statements \cite{cho2018speech}.

For questions, interrogatives which might be represented by \textit{do} support or \textit{wh-} movement in English, were primarily considered\footnote{Note that this does not always hold for the Korean language, which is \textit{wh-in-situ}. A more complicated and audio-aided identification is required in those cases, as in \citet{cho2019disambiguating}}. The ones in an embedded form were also counted, possibly with the predicates such as \textit{wonder}. Also, a large number of the declarative questions \cite{gunlogson2002declarative} were taken into account. Since the corpus utilized in both \citet{cho2018speech} and this annotation process does not contain punctuation marks, the final work was carried out for clear-cut questions that were selected upon the majority voting of the annotators, at the same time removing the utterances that depend on acoustic features. For all the types of questions, the ones in rhetorical tone were removed since their argument usually does not perform as an effective question set \cite{rohde2006rhetorical}.

For commands, the imperatives in a covert subject and with the modal verbs such as \textit{should} were primarily counted. The requests in question form were also taken into account. All the types incorporate the prohibition. Conditionalized imperatives were considered as command only if the conditional junction does not negate the to-do-list. Same as the former case, the ones in rhetorical tone or usage were removed despite it has an imperative structure  \cite{han2000structure,kaufmann2016fine}. All the other types of utterances except questions and commands were considered non-directive\footnote{We aim to explain the type of utterances which are also counted as non-directive in other languages, even if a 1:1 mapping might not be possible through translation. We plan to publish an expansion of this work, which is specific to English sentences accompanied by sample corpora as separate work.}.

\subsection{Annotating Intent Arguments}

The following section exhibits example annotation of intent arguments for non-canonical directives, as shown in Figure 2. We want to note again that while we describe the procedure based on simplified English sentence examples, the actual data and process were significantly more complicated.

\subsubsection{Questions}

For the three major question types, which are defined as \textit{yes/no}, \textit{alternative} and \textit{wh-}\footnote{Note that here, these are not the syntactic properties, but the level of speech act.}, we applied different extraction rules. For \textit{yes/no} questions (\textit{yes/no} Q), we employ an \textit{if-} clause which constraints the candidate answers to yes or no (3a). For \textit{alternative} questions (Alt. Q), we employ a \textit{whether - or - } clause accompanied by the list of possible answers (3b). For \textit{wh-} questions (\textit{wh-} Q), the extraction process starts with a lexicon which corresponds with the \textit{wh-} particle that is displayed (3c-d). It is notable that some alternative questions also show the format that is close to the \textit{wh-}questions, with possibly \textit{between} that corresponds with \textit{whether - or -} (3e).\medskip\\
(3) \textit{a. did I ever tell you about how}\smallskip\\
\phantom{(3) } $\rightarrow$ \textbf{if} the speaker told the addressee about how\smallskip\\
\phantom{(3)} \textit{b. you hungry or thirsty  or both}   \smallskip\\
\phantom{(3) } $\rightarrow$ \textbf{whether} the addressee is hungry \textbf{or} thirsty\smallskip\\
\phantom{(3)} \textit{c. how many points you got}\smallskip\\
\phantom{(3) } $\rightarrow$ \textbf{the number} of points that the addressee got\smallskip\\
\phantom{(3)} \textit{d. i want to know about treadstone}\smallskip\\
\phantom{(3) } $\rightarrow$ \textbf{the information} about treadstone\smallskip\\
\phantom{(3)} \textit{e. you know which is hotter in hawaii or guam}\smallskip\\
\phantom{(3) } $\rightarrow$ \textbf{the place} hotter \textbf{between} hawaii \textit{and} guam\medskip

\subsubsection{Commands}

Since the main intent of the commands is analogous to a to-do-list \cite{portner2004semantics}, we annotated an action which the addressee may take, in a structured format. All of these lists start with \textit{to} indeterminate (4a, REQ, requirement), with possibly \textit{not to} for the prohibitions (4b, PH). During this process, non-content-related lexicons such as politeness strategies (e.g., \textit{please}) were not considered in the extraction (4c). %Multiple arguments were concatenated via \textit{and} + \textit{(not) to} (4d).
\medskip\\
(4) \textit{a. i suggest that you ask your wife} \smallskip\\
\phantom{(3) } $\rightarrow$ \textbf{to} ask one's wife\smallskip\\
\phantom{(3)} \textit{b. yeah  but don't pick me up}   \smallskip\\
\phantom{(3) } $\rightarrow$ \textbf{not to} pick the speaker up\smallskip\\
\phantom{(3)} \textit{c. please don't tell my daddy}\smallskip\\
\phantom{(3) } $\rightarrow$ \textbf{not to} tell the speaker's daddy\medskip%\smallskip\\
%\phantom{(3)} \textit{d. you better get back and monitor the regulatory unit}\smallskip\\
%\phantom{(3) } $\rightarrow$ \textbf{to} get back \textbf{and to} monitor the regulatory unit\medskip

\subsubsection{Phrase Structure}

As discussed above, the argument of the questions are transformed into \textit{if-} clause, \textit{whether-} clause or \textit{the-} phrase. Following this logic, the commands are rewritten to either a \textit{to-}clause or \textit{not to-}clause. Except for the \textit{wh-} questions and some alternative questions, all the rewritten sentences contain at least one predicate (verb). Here, note that unlike the English examples displayed above, in the Korean samples, the components that decide the phrase structure (e.g., \textit{if-, whether-, (not) to-}) are all placed at the end of the sentence, mainly due to head-finality. This is to be further described.

\subsubsection{Coreference}

Coreference is a critical issue when extracting the information from the text. It appears a lot in conversational utterances, in the form of pronouns or anaphora. In the annotation process, we decided to preserve such lexicons except for \textit{I}/\textit{we} and \textit{you} since they are participants in the dialog. The concepts which correspond with the two were replaced with either \textit{the speaker(s)} or \textit{the addressee} as shown in (3a-c) and (4b-c); and in some cases with \textit{one(self)} to make it sound more natural (4a).

\subsubsection{Spatial-Temporal and Subjective Factors}

Unlike other question or command corpora, the proposed scheme includes content which requires an understanding of spatial (5a) and temporal (5b) dependencies, namely deixis. These factors are related to the coreference in the previous section, in particular, involving lexicons such as \textit{there} and \textit{then}. Also, the dialog being non-task-oriented results in the content unintentionally incorporating the subjective information, such as current thoughts of the speaker or the addressee. The proposed scheme tries not to ignore such factors in the intent argument (5c-d), to ensure that the core content is preserved.\medskip\\
(5) \textit{a. put your right foot there}  \smallskip\\
\phantom{(3) } $\rightarrow$ to put the right foot \textbf{there}\smallskip\\
\phantom{(3)} \textit{b. i  i don't want to see you tomorrow}   \smallskip\\
\phantom{(3) } $\rightarrow$ not to meet \textbf{tomorrow} \smallskip\\
\phantom{(3)} \textit{c. any ideas about the colour} \smallskip\\
\phantom{(3) } $\rightarrow$ \textbf{the idea} about the colour\smallskip\\
\phantom{(3)} \textit{d. you ought to know what our chances are}\smallskip\\
\phantom{(3) } $\rightarrow$ \textbf{to be aware} about the speaker's chances\medskip

\section{Dataset Construction}

\subsection{Corpus Annotation}

For the argument annotation process, we adopted the corpus constructed in \citet{cho2018speech}, a Korean single utterance corpus for identifying directives/non-directives that contains a wide variety of non-canonical directives. About 30K directive utterances were adopted for the creation of their intent arguments, which are labeled either question or command. The broader categorization on whether the utterance is question or command had been done with moderate agreement $\kappa$ = 0.85 \cite{fleiss1971measuring}, thus, we only annotated the NL queries, simultaneously tagging the subcategories that directly follow the query. The additional tagging and annotation were done by three Korean natives with a background in computational linguistics, and the cross-checking was done with discussion and modification on the conflicts (improper summarization). In detail, the draft query generation was done by two of the annotators, where they cross-checked the work of each. The last annotator thoroughly checked the validity and appropriateness, so that the consensus can be attained from at least three speakers. The detail on this process with the Korean examples is available in \citet{cho2018extracting}. %The basic scheme is provided in Table 1.

We want to emphasize here that our work is not precisely an \textit{annotation task}, but closer to a \textit{rewriting task} with lax constraints on the expected answer. Although the written NL queries  may not be identical for all the same kind of utterances, we hypothesize that there is a plausible semantic boundary for each utterance.

Although our examples are in English, this kind of rewriting supports that the natural language-formatted intent argument can be robust in preserving the purpose of input directives, although the cultural factors such as politeness might influence. We claim that the constraints of our method guarantee this, as we utilize the nominalized and structured terms. Specific considerations when creating a Korean dataset are discussed below.

\paragraph{Head-finality} In the Korean language, due to the head-finality, all of the structured expressions which are used to construct the phrase structure (Section 3.2.3.) goes to the end of the intent arguments. However, in a cross-linguistic perspective, this may not necessarily change the role of the intent arguments. For example, in the Korean sentence \textit{SENT} = ``\textit{mwe ha-ko siph-ni} (what do you want to do)'', which has an intent argument \textit{ARG} = `\textit{cheng-ca-ka ha-ko siph-un kes} (the thing that the addressee wants to do)', the original \textit{SENT} can be rewritten as \textit{SENT*} = ``ARG-\textit{i mwu-ess-ip-ni-kka}''. Here, \textit{SENT*} can be interpreted as ``\textit{what is ARG}'' or ``\textit{tell me about ARG}'', where the core content \textit{ARG} is not lost in translation. % Though displayed merely for a pair of languages, this kind of rewriting supports that the natural language-formatted intent argument can be robust in preserving the purpose of input directives. We claim that the constraints of our method guarantees this, as it utilizes the nominalized and structured terms. While it is difficult to prove that this holds for all possible languages or language pairs, we at least expect this assumption holds for head-first and head-final languages.

\paragraph{Strong Requirements} The term \textit{strong requirement} is not an official academic term, but was coined and proposed for their existence in the corpus. Simply explained, this can be described as a co-existence of a prohibitive  expression (PH) and the canonical requirement (REQ), as we can see in the sentence ``\textit{don't go outside, just stay in the house}''. Even when the prohibitive expression comes immediately before the requirement, such forbidding expressions are not considered as the core content in the final sentence. That is, in these cases, simply understanding it as ``\textit{just stay in the house}'' does not harm the process of query extraction that results in the ideal final form: `\textit{to stay in the house}'. In Korean where scrambling is common, both [PH+REQ] and [REQ+PH] can be valid expressions. In our work, we did not encounter cases where scrambling leads the interpretation of the utterance to be a prohibition.

\paragraph{Speaker/addressee Notation} We consider the notation of coreference crucial in this work. A subject drop is a typical pattern that can be observed in casual spoken Korean. This is different from English, where the agent and the experiencer are explicit. In Korean, they can be dropped and are resolved with context or prosody. Thus, to minimize the ambiguity, we created two separate corpora; one with the speaker/addressee notation, and the other with this information removed. In the former, we classify all possible cases into one of the following five categories: only the speaker (\textit{hwa-ca}), only the addressee (\textit{cheng-ca}), both (\textit{hwa-ca-wa cheng-ca}), none, and unknown. We believe this kind of information will be beneficial for both contextual disambiguation and further research. On the other hand, in the latter, while the specification must be inferred from the context, the output will be closer to what one would encounter in everyday life.

\subsection{Corpus Augmentation}

\subsubsection{What Should be Supplemented}

Above, we used an existing dataset to annotate intent arguments for questions and command utterances, but encountered an imbalance in the dataset - specifically not having enough data for some utterance types, namely Alt. Q, PH, and Str. REQ. Additionally, we concluded that the amount of parallel data was not large enough for the \textit{wh-} question to be useful in real life, taking into account that the extraction of queries from \textit{wh-} questions involves the abstraction of the \textit{wh-}related concept (e.g., ‘destination’ from \textit{where-to-}). To address the issues, we expanded the dataset size by obtaining various types of sentences from intent arguments, specifically via human-aided sentence generation.

\paragraph{Data Imbalance} First, for Alt. Q, PH, and Str. REQ, we needed to ensure the class balance for each utterance type, or at least a sufficient number for the automation. To this end, we manually wrote 400 intent arguments for each of the three types. Specifically, sentences were created at ratio (1 : 1: 1: 1: 4) for \textit{mail, schedule, smart home, weather}, and \textit{other free topics}\footnote{Other topics include the ones that are not mentioned previously, e.g., game, politics, commercials.}, which are considered as usual topics of interest in intelligent agents and also follow the original corpus.

\paragraph{\textit{Wh-} Questions} To enforce the second goal, the supplement of  \textit{wh-}questions, 800 intent arguments were newly written. The topics of each sentence considered in this process are identical to the above. However, the use of \textit{wh}-particles that can hinder the transformations between \textit{wh-}particles and \textit{wh}-related terms was not allowed. This means that the intent arguments were created in the way in which they only expose the nominalized format, and not the \textit{wh-}particles, e.g., \textit{the weather of tomorrow} rather than \textit{what the weather is like tomorrow}. This trend was also applied when constructing additional phrases for some alternative questions above.

\subsubsection{Method and Outcome}

We recruited eight Seoul Korean natives, with diverse academic backgrounds and sufficient knowledge in Korean grammar, to generate the directive sentences from the queries. In detail, with the 2,000 NL queries suggested above, created by other four Korean native speakers, we requested the participants to write ten utterances per phrase as diversely as possible. The guideline was provided to encourage the use of politeness expressions, scrambling, word replacement, etc., for the diversity of expression, and the process was undergone with free QA hours. The output was cross-checked as in the annotation process and was finally augmented to the corpus. The detailed guideline is demonstrated in \citet{cho2020discourse}.
%In this case, the following was recommended:
%
%\begin{itemize}[noitemsep]
%\item Ten sentences should be written in different styles as much as possible. At this time, the style incorporates all of politeness, honorific, and nuance.
%\item You do not have to repeat the terms in the argument, but you can put different words phrases/idioms for the circumstance. The expression should be suitable for spoken language.
%\item It is also recommended to pursue the diversity of sentence forms through scrambling.
%\item In the case of \textit{wh}-question, \textit{wh}-particles are essential, and alternative questions may be inserted in some cases. Both utterance types need not be written interrogatively.
%\item In the case of prohibition, it should be an utterance that does not do any action that the addressee may do, and should have more force than \textit{permission for not doing}. If prohibiting the action is substantially equivalent to requiring another action, replacing it with the expression is not a problem.
%\item Both prohibition and strong requirements need not be imperative but should have the purpose of preventing or forcing the action of the addressee. An active request is also available.
%\item For arguments that include a speaker/addressee, each uses a corresponding pronoun expression. This constructs both corpora with and without speaker/addressee expression.
%\end{itemize}
%The sentence generation process given the above guideline resulted in a total of

The paraphrasing process resulted in a total of
20,000 argument-directive pairs, constructed from 2,000 arguments. Examples of various question and command expressions for phrases obtained in this process include, for example (from \citet{cho2020discourse}),\medskip\\
\textbf{Argument}: The most important concept in algebra  \\
\textbf{Topic}: Free, \textbf{Type}: \textit{wh-} question  \smallskip\\
$\rightarrow$ \textit{just pick me one important concept in algebra}\smallskip\\
$\rightarrow$ \textit{what you think the core concept in algebra is} \smallskip\\
$\rightarrow$ \textit{which concept is the most important in algebra} \smallskip\\
$\rightarrow$ \textit{what should i remember among various concepts in algebra} $\cdots$  (various versions in Korean)\medskip\\
The composition of the entire dataset after augmentation is shown in Table 1. %We ensured the ratio between the utterance types to be balanced, so that utterances which were not statistically well-represented in the corpus fulfill enough training samples. Additionally, we increased the absolute count of utterances for \textit{wh-} Q where our approach can be proven most effective. 
As a result of the above remedies, the class imbalance and practicability, which were problematic at the initial point, have been partially resolved. The details are available online\footnote{\url{https://github.com/warnikchow/sae4k}}.

\begin{table}[]
	\centering
	\resizebox{\columnwidth}{!}{%
		\begin{tabular}{|c|c|c|c|c|}
			\hline
			\textbf{Intention}                 & \textbf{Types} & \textbf{Original} & \textbf{Augmented} & \textbf{Sum}    \\ \hline
			\multirow{3}{*}{\textbf{Question}} & Yes/no Q       & 5,715             & -                  & 5,715           \\ \cline{2-5} 
			& Alternative Q  & 229               & 4,000              & 4,229           \\ \cline{2-5} 
			& Wh- Q          & 11,988            & 8,000              & 19,988          \\ \hline
			\multirow{3}{*}{\textbf{Command}}  & Prohibition    & 478               & 4,000              & 4,478           \\ \cline{2-5} 
			& Requirement    & 12,302            & -                  & 12,302          \\ \cline{2-5} 
			& Strong REQ.    & 125               & 4,000              & 4,125           \\ \hline
			& \textbf{Total} & \textbf{30,837}   & \textbf{20,000}    & \textbf{50,837} \\ \hline
		\end{tabular}%
	}
	\caption{The final composition of the dataset.}
	\label{tab:my-table}
\end{table}

%\subsection{Dataset}
%In the labeling and annotating process, we've utilized Cornell movie dialog corpus \cite{danescu2011chameleons} which contains a bunch of non-canonical directives. Due to being highly scripted, it also contains many descriptive and rhetorical sentences. Thus, we've tagged the utterances whether those belong to questions, commands, or none of them. The tagging was performed only for the lines that have more than 15 characters since the portion of fragments or simple greetings dominates for the shorter lines. 
%
%The number of total utterances reaches around 21K. The annotation for all the utterances was done by two bilinguals and one L1 speaker. The final decisions were done by majority voting. We've obtained 2,710 questions and 2,109 commands. The inter-annotator agreement (IAA) was $\kappa$ = 0.75 \cite{fleiss1971measuring}.

\section{Experiments}

Here, we validate the usefulness of the constructed dataset with multiple sequence-to-sequence (Seq2Seq) \cite{sutskever2014sequence} architectures. We would like to note that as we propose both a new dataset accompanied by a new task, there is no baseline or proven evaluation metric as of the time of writing. For these reasons, we used existing evaluation frameworks used by other generation tasks.

\subsection{Format}

The final format of the corpus is as follows: [\textbf{\textit{Label / Sentence / Argument}}]. Here, the label denotes the six utterance types as in Section 4.1, and the utterance and intent argument are in raw text form. As stated in Section 4.1.2, there are two versions of the corpus: with and without the speaker/addressee notation. The latter is utilized at this phase, to ensure whether the non-functional contents are well captured.

In the automation process, we aimed to infer the intent argument directly, by giving a \textbf{sentence} as an input and an \textbf{argument} as a target. Here, the correct inference of the intent argument is not independent with the identification of the exact utterance type\footnote{Nonetheless, we don’t consider this task as a classification that identifies the label.} due to the formats being distinct. Therefore, we separate metrics for different tasks. We discuss this further in the evaluation section.

\subsection{Automation}

While the total volume is not significant for fluent automation concerning the usual dataset size for machine translation (MT), we proceeded to observe how the proposed scheme works. The implementation was done through a recurrent neural network (RNN)-based encoder-decoder (enc-dec) with attention \cite{cho2014learning,luong2015effective} and a Transformer \cite{vaswani2017attention}. For the agglutinative nature of the Korean language, morpheme-level tokenization was done with MeCab\footnote{\url{https://bitbucket.org/eunjeon/mecab-ko-dic/src/master/}} tokenizer provided by the KoNLPy \cite{park2014konlpy} library.

For the \textit{RNN enc-dec with attention} that utilizes the morpheme sequence of maximum length 25, hidden layer width and dropout rate \citet{srivastava2014dropout} was set to 512 and 0.1, respectively. This model was trained for 100,000 epochs.

For the \textit{Transformer}, which adopts a much more concise model compared to the original paper \cite{vaswani2017attention}, the maximum length of the morpheme sequence was set to also 25, with hidden layer width 512 and dropout rate 0.5. Additionally, multi-head attention heads were set to 4, and a total of two Transformer layers were stacked, considering the size of the training data. Due to the higher computation budget required, this model was trained for 10,000 epochs.

\subsection{Evaluation}

The most challenging part of validating a new dataset and task is deciding a fair and robust evaluation framework. This is particularly challenging for generative tasks, such as translation or summarization. For this kind of task, several candidates exist that can be considered felicitous for an input utterance. It means that the same phrase can be expressed in various ways, without harming the core content.

Nonetheless, as it is for paraphrasing or summarization, we believe that there should be a rough boundary regarding our tolerance of the output variance. Specifically, in our task, the answer \textit{has to be} some formalized expression. However, if we utilize only BLEU \cite{papineni2002bleu} or ROUGE \cite{lin2004rouge} as a measure, there is a chance that the diversity of possible outputs can result in grammatically incorrect or incomprehensible output \cite{matsumaru-etal-2020-improving}, although it is semantically plausible. Also, in the corpus construction, we have explicitly set the formats for different utterance types, which requires the correct identification of the speech act and thus can largely influence the accurate inference of an argument.

In this regard, we first surveyed a proper metric for the automatic and quantitative analysis of the result, respectively. A part of the conclusion is that the automatic analysis of semantic similarity can be executed utilizing the recent pre-trained language model-based scoring system, namely BERTScore\footnote{BERT denotes a bidirectional encoder representation from Transformer \cite{devlin2018bert}, a freely available pre-trained LM, and this fine-tuned evaluating toolkit is provided in \url{https://github.com/Tiiiger/bert\_score}} \cite{zhang2019bertscore}. Such an approach can be adopted regardless of whether the label is correctly inferred and reflects the common sense inherited in the pre-trained language models. Moreover, in  case the label is correct and some format-related tokens (e.g., \textit{the method}, \textit{whether}, \textit{not to}) in the output overlap with the ones in the gold data, the lexical similarity can also be taken into account, probably as an extra point. It can be further represented by ROUGE compared to the gold standard. 

Considering the different natures, we determined to aggregate both kinds of evaluation values. The final score was obtained by averaging those two results, namely ROUGE-1 and BERTScore. With this, we compensate for the case that the format difference caused by the wrong label leads to the misjudgment on lexical features.

\subsection{Result}

The validation results are in Table 2. For \textit{Total}, we averaged BERTScore and ROUGE-1. 

The result shows the advantage coming from (a) adopting the self-attention-based \cite{vaswani2017attention} Seq2Seq  and (b) setting aside a larger volume of data for the training phase. (a) can be observed in the results, in both ROUGE-1 and BERTScore, where the Transformer model performs better with the same split model, even with the 7:3 split model that has gone through less training. (b) is observed within the two Transformer models. The main reason for the difference is assumed to be the existence of out-of-vocabulary (OOV) terms in the test set, which in our experiments loses information during encoding. As the information has been lost, this in turn affects the performance of the decoder.

\begin{table}[]
\centering
\resizebox{0.8\columnwidth}{!}{%
\begin{tabular}{|c|c|c|c|}
\hline
\textbf{}           & \textbf{\begin{tabular}[c]{@{}c@{}}RNN S2S\\ + Attention\end{tabular}} & \multicolumn{2}{c|}{\textbf{Transformer}} \\ \hline
\textbf{Test split} & 9:1                                                                    & 7:3                 & 9:1                 \\ \hline
\textbf{Iteration}  & 100,000                                                                & 10,000              & 10,000              \\ \hline
\textbf{ROUGE-1}    & 0.5335                                                                 & 0.5383              & 0.5732              \\ \hline
\textbf{BERTScore}  & 0.7693                                                                 & 0.8601              & 0.9724              \\ \hline
\textbf{Total}      & \textbf{0.6514}                                                        & \textbf{0.6992}     & \textbf{0.7728}     \\ \hline
\end{tabular}%
}
\caption{Validation result with the test set.}
\label{tab:my-table}
\end{table}

Beyond the quantitative analysis that mainly concerns metrics, we checked the model's validity with the output for a test utterance that is fed as a common input. For example, from the original sentence (Str. REQ):\medskip\\
(6) ``저번처럼 가지 말고 백화점 세일은 미리 가서 대기하렴'' / ``\textit{This time, please go to the department store earlier (than its opening time) and wait there for the upcoming sale event}''\medskip\\
the followings are obtained from each model:\medskip\\
(6) a. \textbf{RNN Seq2Seq with attention} - 백화점 가 미리 가 서 대기 대기 대기 ... / \textit{department store, go earlier (than its opening time), and wait wait wait ...} (failure of proper phrase ending)\smallskip\\
\phantom{(3) }b. \textbf{Transformer (split 7:3)} - 백화점 가 서 미리 가 서 도와 주 기 / \textit{to go to the department store earlier (than its opening time) and help (something)}\smallskip\\
\phantom{(3) }c. \textbf{Transformer (split 9:1)} - 백화점 세일 은 미리 가 서 대기 하 기 / \textit{to go to the depratment store earlier (than its opening time) and wait for the sale event}\smallskip\\
Taking into account that the given utterance (6) is a strong requirement, or a series of (less meaningful) PH and (substantial) REQ, it is encouraging that all three models succeeded to place the \textit{department store} (백화점, payk-hwa-cem) at the very first of the sentence, ignoring the PH in the first half clause\footnote{Though omitted for the fluent translation, ‘저번처럼 가지 말고’ is PH that originally means \textit{not to go as the last time}.}. However, note that in (6a), the hallucination took place in the RNN model, while the other two Transformer models cope with it and find the right place to finish the inference. Being able to determine when to terminate the sequence is important for matching the sentence type correctly, especially in a head-final language as Korean\footnote{Stably guessing the accurate tail of the phrase is not guaranteed in the auto-regressive inference.}. %More sample outputs are available in Appendix A12 .

Besides, comparing (6b) and (6c), where the tails of the clauses (regarding sentence type) were correctly inferred, the latter fails to choose the lexicon regarding  \textit{wait}, instead picking up \textit{help} that may have been trained in a strong correlation with the terms such as \textit{go earlier} in the training phase. Here, it is also assumed that loanwords such as \textit{sale} (세일, seyil), which is expected to be OOV in the test phase, might have caused the failure in (6b), even though it exists in the input sentence. %The comparison with the proposed measure and the implemented models convey that our approach can have a further development.
The gold standard for (6) is `백화점 세일은 미리 가서 대기하기, \textit{to go to the department store earlier and wait for the sale event}', which is identical to (6c) if the morphemes are accurately merged. %This suggests that the self attention-based model architecture and the supplement of the dataset are both the solution for the stable inference. 

Here are more samples that come from the well-performing Transformer model, especially some tricky input sentences (7) and \textit{wh-} questions (8). We expect such formalizations can be meaningful for the future AIs with personality, aiming at human-friendly interaction. As part of the pre-processing pipeline, all punctuation was removed from the input, and the output phrases were not polished to deliver the original format. \medskip\\
(7) ``박사 졸업과 결혼 준비를 비교한다면 어떤게 더 지옥같아'' / ``\textit{which is more hell if you compare your phd with your wedding preparation}''\smallskip\\
$\rightarrow$ 박사 졸업 과 결혼 준비 중 더 힘들 었 던 것   / \textit{the tougher process (for the addressee) between getting phd and preparing wedding}\medskip\\
(8) ``몇 도 기준으로 열대야라고 해'' / ``\textit{from what temperature is it called a tropical night}''\smallskip\\
$\rightarrow$ 열대야 기준 온도 / \textit{the reference temperature of tropical night}

\subsection{Discussion}

\paragraph{Analysis} The results suggest that the self attention-based model architecture can be quite beneficial for stable inference. Moreover, the inference seems to take advantage of grasping the proper interaction between long-distance components of the input sentence. It emphasizes that the intent argument extraction requires the understanding beyond the given lexicons, not merely being a syntactic parsing task. 
%\paragraph{Validity} The automation of the proposed extraction scheme aims at checking the real utility of the corpus.
Though we cannot rule out the possibility of overfitting, Seq2Seq-style approaches are validated with a moderate amount of sentence-query pairs (40K - 50K). The overall performance is expected to boost up with the modern noise-robust sentence encoders \cite{lewis2019bart}.

\paragraph{Limitation} As shown by the dependency on the train set data, domain generalization issues regarding OOVs is critical in coping with the resource shortage and guaranteeing efficiency. However, we assume that our limitation in the topic may not affect much on generalization given the controllable and content-preserving technologies \cite{logeswaran2018content,martin2019controllable}, since our transformation rarely changes the domain-specific contents. For instance, “\textit{what will you best recommend memorizing in the algebra textbook}” is transformed to `\textit{the most important concept in the algebra}', where the transformation engages in the general expressions (\textit{best, recommend, most important}). That is, though our baseline experimental results merely attest to the validity of the corpus, we believe that models that have higher robustness to OOV, such as those pre-trained on large corpora, will perform better and leverage our framework.

\paragraph{Application} Since the task domain of the proposed approach is not specified, we expect our scheme and output to be worthwhile for a general AI that aims human-friendliness. At the same time, it may prevent users from feeling isolated by talking mechanically. Also, along with the non-task-oriented dialogues, our scheme may be useful for avoiding inadvertent ignorance of the users’ will, such as the digitally marginalized.

\section{Conclusion}

The significance of this research is in proposing the construction and augmentation schemes for rewriting of less explored sentence units, and making it an open, permissive resource for the general public. The sentence set consists of directive utterances in the Korean language, where the morpho-syntactic property often provide difficulties in information retrieval. Additionally, we propose baselines for the constructed dataset using multiple Seq2Seq architectures, exhibiting that our methodology is practically meaningful in real-world applications. 

Our next work is to extend this more typologically by showing that the annotation/generation scheme applies to other languages. While the scope of our work is limited to Korean, we hope that the proposed annotation scheme and resources from our work can be reused as a common protocol for intent-argument extraction tasks in other languages.

\section*{Acknowledgments}

This research was supported by the Technology Innovation Program (10076583, Development of free-running speech recognition technologies for embedded robot system) funded By the Ministry of Trade, Industry \& Energy (MOTIE, Korea). %Besides, the corpus construction would not have been possible without the help of eight great participants, namely Eunah Koh, Kyung Seo Ki, Sang Hyun Kim, Kimin Ryu, Dongho Lee, Yoon Kyung Lee, Minhwa Chung, and Ye Seul Jung. 
Also, the authors appreciate Siyeon Natalie Park for suggesting a great idea for the title. %We also thank the anonymous reviewers for the constructive suggestions. 
After all, the authors are grateful for the invaluable advices and supports provided by Reinald Kim Amplayo, David Mortensen, Jong In Kim, Jio Chung, and $\dagger$Kyuwhan Lee.

% \nocite{*}
%\section{Bibliographical References}
%\label{main:ref}

\bibliographystyle{acl_natbib}
\bibliography{my_bib_warnik}

%\section{Language Resource References}
%\label{lr:ref}
%\bibliographystylelanguageresource{lrec}
%\bibliographylanguageresource{lrec2020W-xample}

\end{document}